\documentclass[10pt, conference, compsocconf]{IEEEtran}
\usepackage{cite}

\ifCLASSINFOpdf
  \usepackage[pdftex]{graphicx}
\else
\fi
\usepackage[cmex10]{amsmath}

\usepackage{algorithm}
\usepackage{algpseudocode}
\usepackage{amssymb}
\usepackage{paralist}

\usepackage{multicol}
\usepackage{multirow}
\usepackage{subcaption}

\DeclareMathOperator*{\argmax}{arg\,max}
\DeclareMathOperator*{\argmin}{arg\,min}

\renewenvironment{description}[1][0pt]
  {\list{}{\labelwidth=0pt \leftmargin=#1
   }}
  {\endlist}

\begin{document}
%
\title{\Large{\textbf{Inferring Restaurant Styles by Mining Crowd Sourced Photos \\ from
User-Review Websites}}}


\author{\IEEEauthorblockN{Haofu Liao, Yuncheng Li, Tianran Hu, Jiebo Luo}
\IEEEauthorblockA{Department of Computer Science\\
University of Rochester\\
Rochester, New York 14627, USA\\
\{hliao6, yli, thu, jluo\}@cs.rochester.edu}
}


%


\maketitle

\begin{abstract}
When looking for a restaurant online, user uploaded photos often give people an immediate and tangible impression about a restaurant. Due to  their informativeness, such user contributed photos are leveraged by restaurant  review websites to provide their users an intuitive and effective search  experience. In this paper, we present a novel approach to inferring restaurant  types or styles (ambiance, dish styles, suitability for different occasions)  from user uploaded photos on user-review websites. To that end, we first collect a novel restaurant photo dataset associating the user contributed photos with the  restaurant styles from TripAdvior. We then propose a deep multi-instance multi-label learning (MIML) framework to deal with the unique problem setting of the restaurant style classification task. We employ a two-step bootstrap strategy to train a multi-label convolutional neural network (CNN). The multi-label CNN is then used to compute the confidence scores of restaurant styles for all the images associated with a restaurant. The computed confidence scores are further used to train a final binary classifier for each restaurant style tag. Upon training, the styles of a restaurant can be profiled by analyzing restaurant photos with the trained multi-label CNN and SVM models. Experimental evaluation has demonstrated that our crowd sourcing-based approach can effectively infer the restaurant style when there are a sufficient number of user uploaded photos for a given restaurant.

\end{abstract}


%
\IEEEpeerreviewmaketitle

\section{Introduction} \label{sec: introduction}

Nowadays, more and more people rely on user-review websites, such as
Foursquare, Yelp, and TripAdvisor, to find a restaurant, hotel or other venues. Searching restaurants online
is fast and convenient. By a single tap on the screen, people can easily find
thousands of restaurants in their cities. For most of the recommended restaurants,
the user-review websites will usually provide much basic information,
such as the price range, location, operation hours, to their users. Apart from
the basic information provided by the websites, there are
usually many user contributed photos. People can therefore get a more intuitive
impression of the restaurant by looking at the pictures taken by other users.

\begin{figure}
\centering
\includegraphics[scale=0.18]{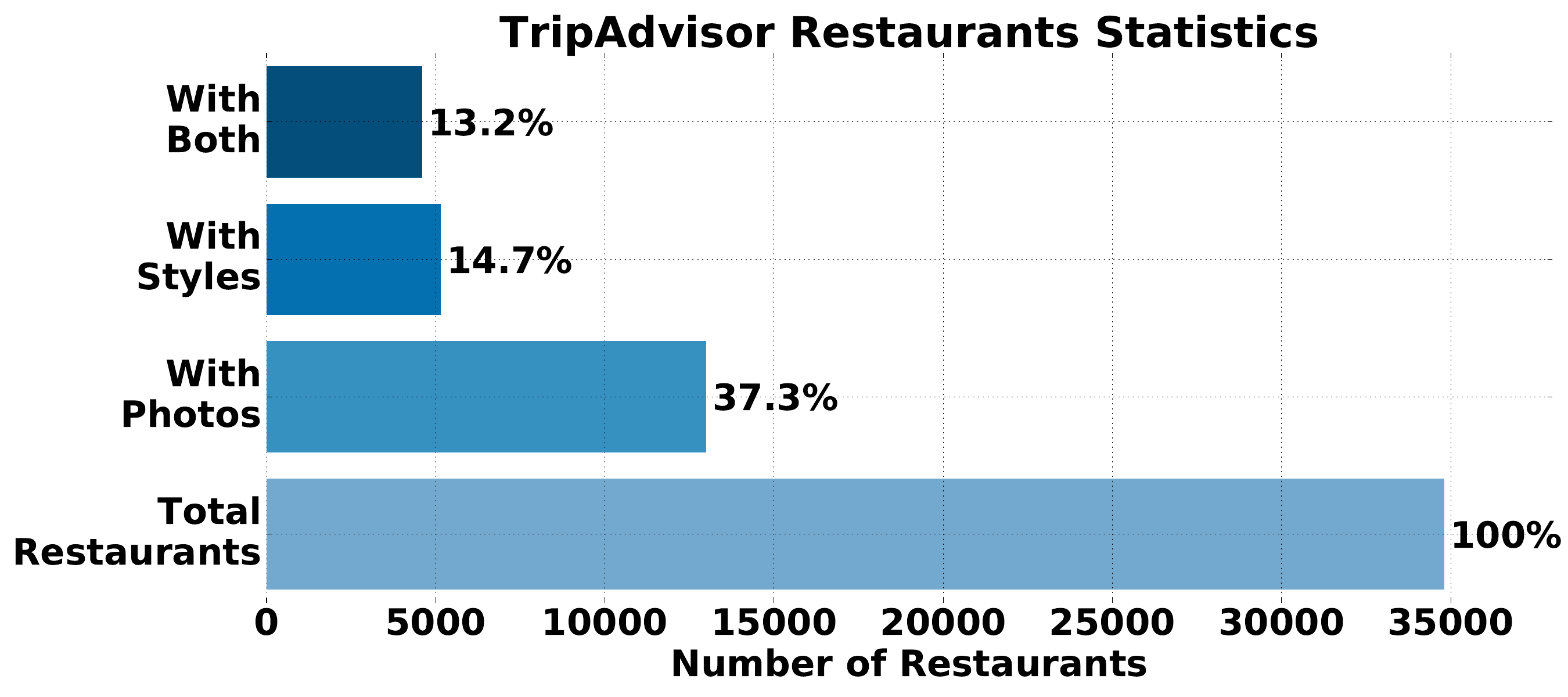}
\caption{Statistics of restaurants on TripAdvisor.}
\label{fig: stats}
\end{figure}

However, the basic information and user contributed photos of restaurants cannot
satisfy all the needs from users. Consider the situation that someone wants to celebrate
a wedding anniversary with his wife and has no idea which restaurant is the
best. Or he needs to schedule a dinner meeting with his clients and wonders
which restaurant is conducive for a business conversation. Such demands
from users require higher level information of the restaurant and they are not
directly available from the restaurant's basic information. In this paper, we refer to such higher level information as the \textit{style of restaurant}. Currently, user-review websites either contain
no restaurant styles, such as Foursquare, or rely on users feedbacks, such as
TripAdvisor. Therefore, for most of the restaurants online, such information is not
available. As shown in Figure \ref{fig: stats}, among the $34,787$ restaurants
 we extracted from TripAdvisor, only $14.7\%$ of them have style tags.
In this paper, we will refer to styles and tags interchangeably.
 A even smaller percentage of restaurants has user contributed photos to
support the style tags. Only $13.2\%$ of the collected restaurants
data contain both photos and style tags.

Although the style tags are not commonly available from the user-review
websites, fortunately, it is possible that they can be derived from the user
contributed photos \footnote{Note that there might be some other information exists to help inferring
restaurant style, such as the text reviews. However, in this work, we will only focus on classifying restaurant styles using user-uploaded photos}, which have a much higher availability. For example, if we
see photos from a restaurant containing delicate dishes and romantic decorations,
we can infer that this restaurant is suitable for a wedding anniversary or Valentine's Day dinner.
Hence, we propose a restaurant style inference method based on the analysis
of the user contributed photos in user-review websites. There are multiple applications for the proposed system:
\begin{inparaenum}[1)]
  \item For a given restaurant, it can estimate a collection of highly related restaurant styles from user contributed photos;
  \item It can then be used to guide user-review sites to present more useful information of restaurants. The additional availability of the restaurant style tags can help users to search particular types of restaurants;
  \item For a restaurant without photos, our method can be used to present a collection of
    strongly related images that match the style tags of the restaurant. Thus,
    the users can have a good idea about the style of such a restaurant even without
    seeing the true images of the exact restaurant;
  \item The model can help the restaurants to select the best photos to advertise their restaurant styles; and 
  \item  The trained models can be used by other websites and services related to restaurants.
\end{inparaenum}

Inferring restaurant styles from user-uploaded photos can be best described as
a multi-instance multi-label learning (MIML) problem \cite{NIPS2006_3047} where
each object (i.e., restaurant) is described by a set of instances
(i.e., user-uploaded photos) and associated with several class labels (i.e., restaurant style tags).
Traditional MIML problems usually assume that
\begin{inparaenum}[1)]
  \item each instance contributes equally and independently to the object's class label;
  \item or there exists a ``key'' instance that contributes the object's class label.
\end{inparaenum}
However, for the restaurant style classification problem such assumptions do not
hold. Instead, it is the collection of ``key'' instances that decide the object's
class label. Based on such assumption, we propose a deep MIML approach
to address this problem. We train a multi-label CNN for two rounds in a bootstrap
fashion. We then use the trained multi-label CNN in combining with a restaurant profiling
algorithm to extract restaurant style features from the collection of
images of a restaurant. Next, we feed the extracted features to a set of SVMs to
obtain the restaurant style tags. Our experimental results show that the proposed
method is indeed effective in predicting the style tags of a restaurant and
a F-1 score of 0.58 is achieved.

The remainder of this paper is organized as follows. In Section \ref{sec: related work},
we present on the related work. In Section \ref{sec: data collection}, we introduce
the procedure of data collection and statistics of the collected dataset.
The proposed framework and detailed algorithms used in our method are described in
Section \ref{sec: approach}. We evaluate the performance of our method in
Section \ref{sec: experimental results} and conclusions are given in
Section \ref{sec: conclusions}.

\section{Related Work} \label{sec: related work}

\subsection{Related Work in Social Media}

Mining data from user-review websites is popular in recent years
\cite{6413812,cheng2011exploring,Yuan:2013:TPR:2484028.2484030,Wang:2010:LAR:1835804.1835903,doi:10.1287/mksc.1110.0700,Zahalka:2014:NYM:2647868.2656403}. However, only a few of them focus on data mining on 
restaurants \cite{Mui:2001:CSA:375735.376020,1287319,lee2006location,Liu2001455,fu2014user}.
Aside from our work, which uses visual information, most these restaurant related
works only use restaurant meta data, user reviews, or geographic information
to provide restaurant recommendations to users \cite{lee2006location,1287319,fu2014user} and
none of them is able to provide higher level recommendations, such as restaurant
styles. Other venue related works make use of visual
information to decide which venues should be recommended 
\cite{Zahalka:2014:NYM:2647868.2656403,6403552,Kofler:2011:NAP:2072609.2072624,Cheng:2011:PTR:2072298.2072311}.
In particular, \cite{Zahalka:2014:NYM:2647868.2656403}  uses deep neural
networks to extract useful features in images. What sets our work apart from this work
is that we focus more on the restaurant styles, which is highly abstract, while they mainly concentrate on general and more concrete aspects
of a venue. 

\subsection{Related Work on Multiple Instance Multiple Label Learning}

The formal definition of a MIML framework is given by Zhou et al \cite{NIPS2006_3047}. 
They proposed two ways of targeting the MIML problem:
\begin{inparaenum}[1)]
\item Degenerating the MIML problem to a multi-instance learning problem and solving
the problem with a multi-instance learner;
\item Degenerating the MIML problem to a multi-label learning problem and solving
the problem with a multi-label learner.
\end{inparaenum}
However, existing MIML algorithms \cite{4587384,Zhou20122291,huang2013fast} are
not applicable to our restaurant style classification problem. \cite{Zhou20122291}
assumes that all the instances contribute equally to the object's labels.
However, in our case, there are a number of uninformative images, such as
pictures of common dishes and images of the dinning tables. They contribute
very little to the restaurant's style. \cite{4587384,Zhou20122291}
are only feasible when the numbers of possible instances and objects are not large.
While, in our case, we need to train and test our model on a large-scale dataset
(See Table \ref{tab: city stats}). \cite{huang2013fast} is provably efficient to
work on large datasets. However, it assumes that the object is positive if and only if there
exists at least one positive instance. In our case, such assumption is not
guaranteed. For example, just one picture of a delicate dish does not mean
the restaurant itself is romantic.

\subsection{Related Work on Convolutional Neural Network}

In recent years, deep convolutional neural networks (CNNs) have shown to be very successful
in lots of machine learning tasks. In relating to our problem, some literatures
target the multi-instance learning problems using CNNs. Pathak et al. \cite{DBLP:journals/corr/PathakSLD14} proposed
a novel multi-instance learning algorithm for semantic segmentation using CNNs.
The proposed algorithm try to learn pixel-level semantic segmentation only based
on the weak image-level labels, while in our case we need to learn image-level
labels from bag-level restaurant labels. Wu et al. \cite{7298968} showed that deep
learning based multi-instance learning can be very useful in image classification
and auto-image notation. The authors try to classify images based on a set of generated
image proposals and therefore the proposed algorithm relies heavily on the assumption
that if there exists at least one positive proposal then the image itself is positive.
However, such assumption does not hold for our restaurant classification problem. There are some other
works focus on solving multi-label learning problems using CNNs. Wei et al. \cite{DBLP:journals/corr/WeiXHNDZY14}
proposed a deep CNN infrastructure, called Hypotheses-CNN-Pooling, for multi-label
image classification. Gong et al. \cite{DBLP:journals/corr/GongJLTI13} investigated 
the choices of different loss functions on the performance of multi-label image
annotation using CNNs. However, none of these works consider the MIML problem
using CNNs, which sets our work apart from existing CNNs algorithms.

\section{Data collection} \label{sec: data collection}

\begin{table}
  \centering
  \begin{tabular}{l|c|c}
    \hline
    \textbf{Cities} & \textbf{\# of Restaurants} & \textbf{\# of Photos} \\
    \hline
    Chicago         & 7150                      & 32676                \\
    Houston         & 6732                      & 14084                \\
    Los Angeles     & 7903                      & 22379                \\
    New York City   & 9871                      & 128657               \\
    Miami           & 3131                      & 19672                \\
    Total           & 34787                     & 217468               \\
    \hline
  \end{tabular}
  \caption{Restaurant statistics of five major US cities.}
  \label{tab: city stats}
\end{table}

\begin{table}
  \centering
  \begin{tabular}{cc}
    \hline
    \multicolumn{2}{c}{\textbf{Restaurant Styles From TripAdvisor}} \\
    \hline
    Bar scene                & Local cuisine       \\
    Business meeting         & Romantic            \\
    Dining on a budget       & Scenic view         \\
    Families with children   & Special occasions   \\
    Large groups             & \\
    \hline
  \end{tabular}
  \caption{Restaurant styles in TripAdvisor.}
  \label{tab: restaurant styles}
\end{table}

All of our data are collected from TripAdvisor. We choose TripAdvisor because some of its restaurants contain user
labeled style tags. We can use them as the pseudo ground truth to guide
the training of a multi-label CNN, which will be discussed in Section
\ref{sec: multi-label cnn}. However, as we have mentioned in Section
\ref{sec: introduction}, for most of the restaurants, the style tags are
not available. Therefore, we still need to use the trained framework in Section
\ref{sec: approach} to assign them restaurant styles based on the associated user contributed photos.

To make sure the restaurant photos we use are representative, we choose restaurants
from five major cities in the United States. A list of cities with the number of restaurants
and photos is given in Table \ref{tab: city stats}. In total, our data set
contains $34,787$ restaurants and $217,468$ photos. According to TripAdvisor's
taxonomy, there are in general 9 restaurant styles. The taxonomy of these styles
is shown in Table \ref{tab: restaurant styles}.

\section{Methodology} \label{sec: approach}

\begin{figure}
  \centering
  \includegraphics[scale=0.43]{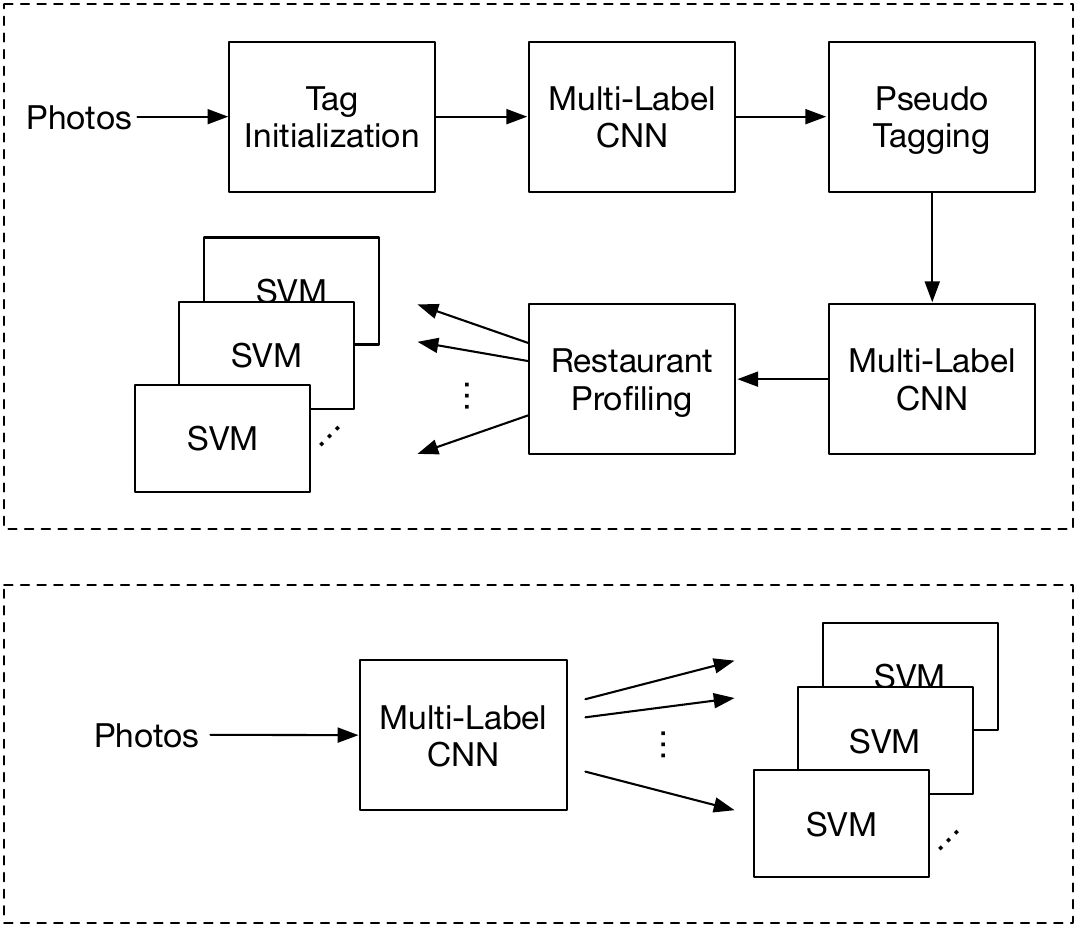}
  \caption{Illustration of the framework of our method. Top: training phase.
  Bottom: testing phase.}
  \label{fig: framework}
\end{figure}

The framework of our approach is illustrated in Figure \ref{fig: framework}.
The flowchart at the top demonstrates the training phase and the bottom represents the
testing phase. For a given restaurant, our goal is to detect a collection of
applicable restaurant styles from user contributed photos. To deal with this
MIML learning task, we adapt the idea from \cite{NIPS2006_3047}: reducing the task to
a multi-label learning task and solving it with a multi-label learner. In a high
level view, our proposed framework extracts the features of a restaurant from
a bag of images using CNNs and then feeds the features to a series of binary SVMs
to obtain the restaurant styles.

In the training phase, we build a multi-label CNN that can infer the style features
represented by the collection of user-contributed photos from a restaurant.
For each photo, the multi-label CNN outputs a vector of scores where
each score indicates the confidence of the photo relating to a restaurant
style. The score vectors will be used to obtain the style features of a restaurant.
However, due to the multi-instance learning nature of our problem, the challenge
we face is that we do not have individually labeled photos for direct training. The only
information we have is the style tags of the restaurants, which are supplied by
users (and can be noisy). To solve this problem, we propose to train the
multi-label CNN for two rounds in a bootstrap fashion. As illustrated in Figure
\ref{fig: framework}, the first round CNN plus the pseudo tagging algorithm is
used to estimate the labels of each image in the restaurant. Then, we train the
second round CNN based on the labeled images and use this CNN to extract restaurant
style features in the test phase. Notice that there is no need to train for
the third round (or even further). As we will then train and test on the same
dataset and there will not be any performance gain using the pseudo
tagging algorithm.

In the first round, we aggressively initialize the tags of each photo according
to its restaurant style tags, i.e., all the images from the same restaurant will
have the same style tags as the restaurant. Note that such assignment of tags may result lots
of false positives and negatives. Based on our observations, most of the false 
positives and negatives relate to uninformative images. They frequently appear
across restaurants and contain no information that indicate specific styles.
Those uninformative images are not considered as noise for the multi-label CNN,
as their contribution to each restaurant styles will be canceled out during the
optimization process of the training. In the cases that false positives
and negatives are actually informative images, the multi-label CNN
training may be degraded by those images. However, such cases are not frequent,
otherwise the tags of the corresponding restaurants would be changed. Therefore, in general the
assignments of photos with their restaurant's tags are reasonable and the trained
multi-label CNN classifier should give us good estimates of labels of individual images.
The output of our multi-label CNN is a $9$-dimension score vector, where
each score represents the weight of the image on the corresponding tag. We will discuss
more details of the structure of the multi-label CNN model in Section
\ref{sec: multi-label cnn} and a set of top scored photos is shown in
Figure \ref{fig: top scores}.

\begin{figure*}[h!]
  \centering
  \includegraphics[scale=0.35]{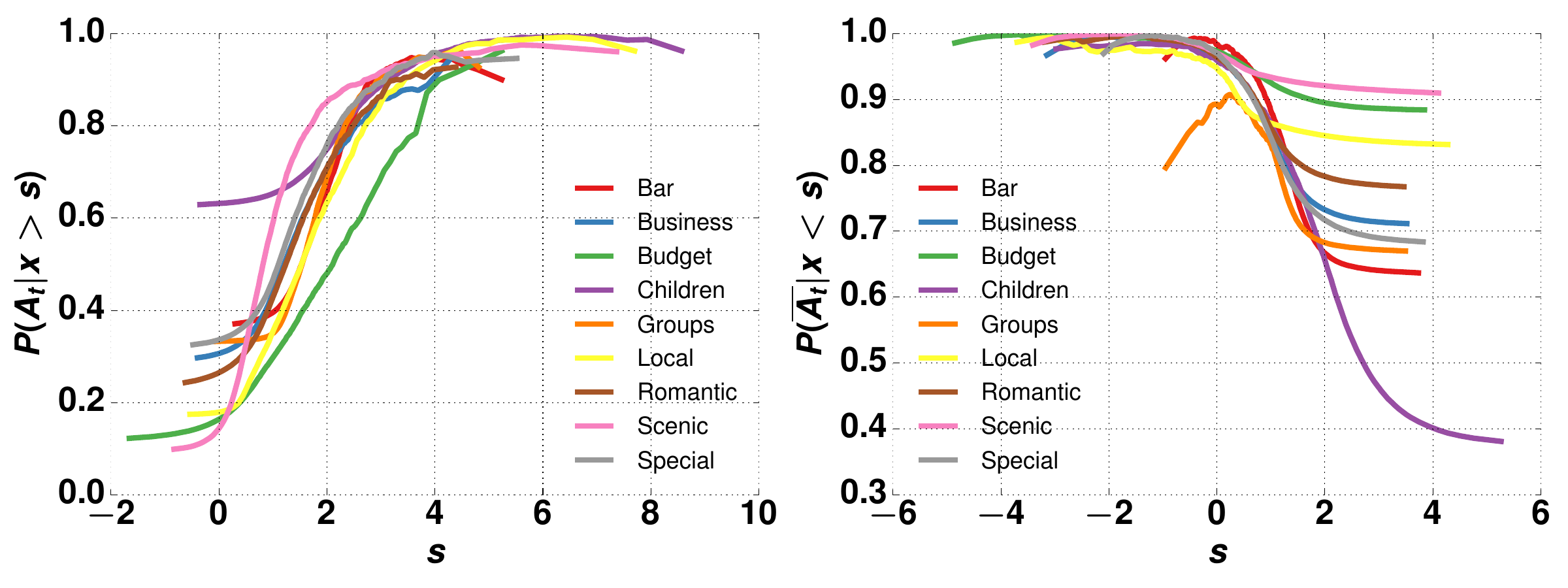}
  \caption{The probability relation between photo scores and restaurant styles.
    Given a photo with a score $x$ on tag $t$ greater than $s$, the probability of its
    restaurant has a style $t$ is denoted by $P(A_t | x > s)$. Given a photo
    with a score $x$ on tag $t$ less than $s$, the probability of its restaurant
    has no style $t$ is denoted by $P(\overline{A_t}|x < s)$. Here, the scores of
    photos are computed from the first round multi-label CNN. $P(A_t | x > s)$
    and $P(\overline{A_t}|x < s)$ for each tag are calculated using Equations (\ref{equ: posterior1})
  and (\ref{equ: posterior2}).}
  \label{fig: posterior}
\end{figure*}

Before training the multi-label CNN for the second round, we use
a pseudo tagging algorithm to relabel the tags of each image in the training set.
Basically, this algorithm takes the scores of each image as an indicator of
being relevant to the tags and remove the presumably false negatives and positives accordingly.
The details of this algorithm are given in section \ref{sec: pseudo tagging}.
There are two reasons that we need this algorithm and the second round
multi-label CNN. First, the pseudo tagging algorithm can remove some incorrectly
labeled informative images and thus can improve the performance of the second round
multi-label CNN. Second, the pseudo tagging algorithm rewards higher scored images
and punishes lower scored images by adding and removing the corresponding tags.
Thus, the outputs of the second round multi-label CNN will be more sensitive to
those images that have higher similarity with the top scored images in the training
set. Such a sensitivity would be helpful to extract better restaurant features
in the restaurant profiling phase. Based on the scores of images, we extract the 
features of each restaurant using the restaurant profiling algorithm, which we
shall discuss more details in Section \ref{sec: restaurant profiling}. We then
train $9$ binary classifiers using SVM, and the outputs of the classifiers give
the tags of the restaurants.

Finally, in the testing phase, we use the entire collection of photos from
a restaurant as the input. Next, compute their scores using the second round
multi-label CNN model. For each label, we extract the restaurant's features
using the top-$k$ scores and use the output of the SVM classifiers to derive
the restaurant styles.

\subsection{Multi-label CNN} \label{sec: multi-label cnn}
\subsubsection{Test}

As we have discussed in the previous section, our approach uses the multi-label CNN
to evaluate each image of a restaurant. The multi-label CNN outputs the score
vectors that will be further used to obtain the style features of a restaurant.
Here, each score indicates the confidence of the photo being related to a
restaurant style.

Formally put, let $\mathbf{X}_i$ and $\mathbf{Y}_i$ denote the images and the associated tags,
respectively. The associated tags $\mathbf{Y}_i = [y_i^1, y_i^2, \dots, y_i^L]^T$
is a binary vector, where
\begin{equation}
y_i^t =
\begin{cases}
1, \text{if tag } t \text{ belongs to } \mathbf{X}_i, \\
0, \text{otherwise.}
\end{cases}
\end{equation}
and $L$ is the total number of tags (in this work $L=9$). Due to its high capacity and
flexibility, CNN has become a common framework to learn image representation
that adapts to specific domains \cite{krizhevsky2012imagenet}. CNN maps images
$\mathbf{X}_i$ to a feature space with multiple layers of convolution, pooling, non linear
activations and fully connected layers. The final layer of the architecture is
designed to output the label confidence scores with respect to the image. Let
$\mathbf{S}_i = [s_i^1, s_i^2, \dots, s_i^L]^T$ denotes the score vector with
respect to $\mathbf{X}_i$. A proper loss function $l(\mathbf{S}_i, \mathbf{Y}_i)$ is designed
over the score vector $\mathbf{S}_i$ to learn the weights $\theta$ in the CNN
architecture using stochastic gradient descent (SGD),
\begin{equation}
  \theta := \theta - \alpha \sum_i \nabla_\theta l(\mathbf{S}_i, \mathbf{Y}_i; \theta),
  \label{eqn:mll-loss}
\end{equation}
where $\theta$ denotes the network parameters to be learned from the training
dataset and $\alpha$ is the learning rate for the optimization algorithm. We use
the cross entropy sigmoid loss function \cite[Chapter~3]{nielsen2015neural} as our form of $l(\mathbf{S}_i, \mathbf{Y}_i)$,
\begin{equation}
  l(\mathbf{S}_i, \mathbf{Y}_i) \triangleq \sum_{t} y_i^t \sigma(s_i^t) + (1-y_i^t)(1-\sigma(s_i^t)),
  \label{eqn:sigmoid}
\end{equation}
where $\sigma$ is the sigmoid function $\sigma(x) = 1/(1+\exp(-x))$ which transforms
label confidence into probabilities. The objective of \eqref{eqn:sigmoid} is to
maximize the confidence of the labels in the target $\mathbf{Y}_i$ and suppress
those not in the target $\mathbf{Y}_i$. From \eqref{eqn:sigmoid}, we compute the
gradients of the loss function with respect to the label confidence $\mathbf{S}_i$,
which is used to compute $\nabla_\theta l(\mathbf{S}_i, \mathbf{Y}_i; \theta)$ for
the stochastic gradient descent algorithm by applying chain rules. This process
is also called \textit{backpropagation}.


\subsection{Pseudo Tagging} \label{sec: pseudo tagging}

As mentioned in Section \ref{sec: approach}, we do not have the ground truth of
the individual training images. Therefore, we use the first round multi-label CNN combined
with the pseudo tagging algorithm to estimate the labels of the training images.
The pseudo tagging algorithm is based on the following assumption:
\begin{itemize}
  \item Images with the highest scores are likely to be related to the tag and images
    with the lowest scores are likely to be unrelated to the tag.
\end{itemize}
This assumption is based on the analysis and observation that the first
round multi-label CNN training in general provides a good indication of being
relevant to the tags. It can be further verified using the following conditional
probabilities. Let $A_t$ be the event that a given style tag $t$ should be assigned
to the restaurant $A$ and $x$ be the confidence score of an image. Thus, given that an
image has the score $x$ greater than $s$, the probability of event $A_t$ can 
be estimated by
\begin{equation} \label{equ: posterior1}
  P(A_t | x > s) = \frac{C(A_t, x > s)}{C(x > s)}.
\end{equation}
Here, $C(A_t, x > s)$ is the number of images whose restaurant tag is $t$ and
have a score on $t$ greater than $s$. $C(x > s)$ is the total number of images
whose score on $t$ is greater than $s$. Note that we do not aggregate images
from the same restaurant yet as we only focus on the score effectiveness of 
individual images. As long as we have the scores of training images from the 
first round multi-label CNN. $C(A_t, x > s)$ and $C(x > s)$ can be easily counted.
Similarly, we can estimate $P(\overline{A_t}|x < s)$ by
\begin{equation} \label{equ: posterior2}
  P(\overline{A_t}|x < s) = \frac{C(\overline{A_t}, x < s)}{C(x < s)}.
\end{equation}
where $\overline{A_t}$ denotes the event that tag $t$ is not assigned to
restaurant $A$. The probability distribution of $P(A_t | x > s)$
and $P(\overline{A_t}|x < s)$ among the training photos is shown in Figure
\ref{fig: posterior}. In the left figure, we can see that in general when
the score of a photo is high on a tag, then it is very possible that its
restaurant will also be assigned to this tag. Similarly, in the right figure,
 those lower scored images on tag $t$ are more likely coming
from restaurants without style $t$. Detailed discussion of this experimental
result is given in Section \ref{sec: perf multi-label cnn}.

Let $\mathbf{X}_{m,n}$ be the $m$th image of the $n$th restaurant in the training set,
$\mathbf{S}_{m,n}=[s_{m,n}^1, s_{m,n}^2, \dots, s_{m,n}^L]^T$ be the score vector with
respect to $\mathbf{X}_{m,n}$, and $\mathbf{Y}_{m,n}=[y_{m,n}^1, y_{m,n}^2, \dots, y_{m,n}^L]^T$ be the
binary label vector associated with $\mathbf{X}_{m,n}$, where
\begin{equation}
y_{m,n}^t = \begin{cases}
1, \text{ if tag } t \text{ belongs to } \mathbf{X}_{m,n}, \\
0, \text{ otherwise.}
\end{cases}
\end{equation}
and $L$ is the total number of possible tags in the dataset.
The score vector $\mathbf{S}_{m,n}$ can be obtained from the multi-label CNN which
we trained in the first round. Since $s_{m,n}^t$ denotes the confidence of
assigning tag $t$ to $\mathbf{X}_{m,n}$ and a higher score denotes a higher confidence, we
can estimate the binary label vector $\mathbf{Y}_{m,n}$ using
$\mathbf{S}_{m,n}$. To this end, we propose the following algorithm:

\begin{description}
  \item[1.] Initialize all the images in the training set with their 
    restaurant tags, i.e., $\forall m,n$, $\mathbf{Y}_{m,n} = \mathbf{Y}_n$, where
    $\mathbf{Y}_n$ denotes the binary label vector of the $n$th restaurant. 
  \item[2.] For each image $\mathbf{X}_{m,n}$ in the training set, compute
  their score vector $\mathbf{S}_{m,n}$ using the multi-label CNN.
  \item[For] each tag $t$ \textbf{do}
  \begin{description}[0.5cm]
  \item[3.] Pick the image $\mathbf{X}_{a,b}$ that has the
    highest score $s_{a,b}^{t}$ among all the images that do not have tag $t$ and
    assign $t$ to $\mathbf{X}_{a,b}$, i.e., pick
    \begin{equation}
      a, b = \argmax_{m, n}\{s_{m,n}^t \mid y_{m,n}^t = 0 \}
    \end{equation}
    and set $y_{a,b}^{t} = 1$.
  \item[4.] Pick the image $\mathbf{X}_{c,d}$ that has the
    lowest score $s_{c,d}^t$ among all the images that have tag $t$ and
    remove $t$ from $\mathbf{X}_{c,d}$, i.e., pick
    \begin{equation}
      c, d = \argmin_{m, n}\{s_{m,n}^t \mid y_{m,n}^t = 1 \}
    \end{equation}
    and set $y_{c,d}^t = 0$.
  \item[5] If $s_{a,b}^t - s_{c,d}^t > const$, go to step $3$.
  \end{description}
  \item[End for]
\end{description}
Here, in step $3$ and $4$, each time we only alter the tag of the image that has
the highest possibility that it was incorrectly labeled. For example, in step
$3$, the algorithm searches the image that does not have tag $t$ but has a
relatively high confidence score on $t$ (the highest among all the
images that are not labeled with $t$). As such an image is very likely to be
incorrectly labeled, we flip its label (from $0$ to $1$). Therefore, in general,
this algorithm will yield a better estimation of the tags in each iteration.

In each iteration of the inner-loop, the confidence scores on tag $t$ of image
$\mathbf{X}_{a,b}$ and $\mathbf{X}_{c,d}$ will keep on decreasing and increasing,
respectively. This is because we always flip the tags of images with highest (in step $3$)
or lowest (in step $4$) confidence scores. When the difference of $s_{a,b}^t$
and $s_{c,d}^t$ becomes very small, the multi-label CNN has a low confidence in
classifying $\mathbf{X}_{a,b}$ and $\mathbf{X}_{c,d}$ correctly. In this case, image
$\mathbf{X}_{a,b}$ and $\mathbf{X}_{c,d}$ are actually uninformative images. Since
the flipping of tags of $\mathbf{X}_{a,b}$ and $\mathbf{X}_{c,d}$ will not help us
distinguish the informative images, we should then stop the label estimation with
respect to tag $t$ and move on to the next tag. Note that as we only want to find the
informative images of each restaurant, those incorrectly labeled uninformative
images will not degrade the performance of the multi-label CNN and that is why
we call this algorithm pseudo tagging. The parameter $const$ in the stop
condition of step $5$ is chosen empirically based on the score distribution of
the uninformative images. The pseudo tagged photos are used to train the
multi-label CNN for the second round. The image scores computed by the second
round CNN will then be used for restaurant profiling.

\begin{figure*}[h!]
  \centering
  \includegraphics[scale=0.15]{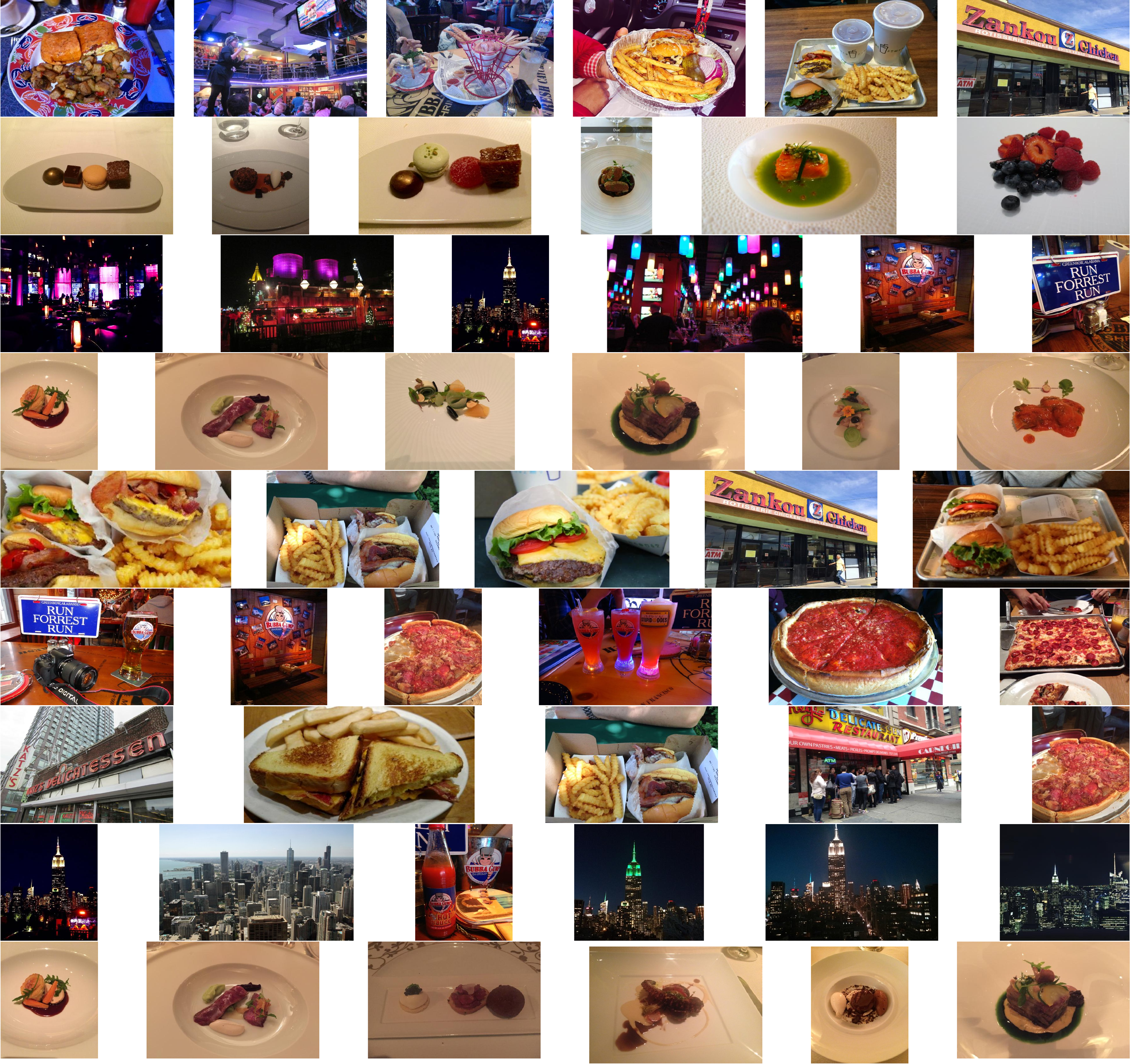}
  \caption{Top scored photos. From top to bottom rows: Family with children, 
  romantic, bar scene, business meetings, dining on a budget, large groups,
  local cuisine, scenic view, special occasions.}
  \label{fig: top scores}
\end{figure*}

\subsection{Restaurant Profiling} \label{sec: restaurant profiling}
To infer the styles of a restaurant, we need to extract the restaurant features
from a collection of photos. However, for a restaurant with a given style,
not all the photos are necessarily related to this style. In fact, in most of the cases,
only a small portion of the images will be useful. For example, for a ``bar scene''
style, we are usually looking for glasses or bottles of wine, bar counters,
and dark scenes with a group of people drinking alcohol. Therefore, images of
regular dishes or foods will not contribute  much to our decision. However, usually
most of the user contributed photos are related to the dishes themselves instead of the
``bar scene''. To this end, we design a restaurant profiling algorithm, which
is based on the following assumption:
\begin{itemize}
  \item The tag of a restaurant is determined by a number of informative images. If a
    restaurant is labeled with a certain tag,  there must be a number of images that are
    highly related to this tag.
\end{itemize}
Based on this assumption, we use the photo scores computed by the
multi-label CNN as an indicator of informativeness. As a higher score means
a higher possibility an image is relevant to a tag, top scored images always
have a better chance that they are the informative images. Thus, we pick the
top-$k$ scored images as the candidates of the informative images and choose their
scores as the extracted features on a given tag. Here, $k$ denotes the number
of the photos we choose to treat as the informative images of a restaurant. We
will discuss more about the choice of $k$ in Section \ref{sec: overall perf}.

Formally put, let $\mathbf{G}_n^t = [s_{1,n}^t, s_{2,n}^t, \dots, s_{M_n,n}^t]^T$
be the score vector of the $n$th restaurant with respect to tag $t$, where
$s_{m,n}^t$, $m\in\{1,\dots,M_n\}$, denotes the confidence score of image
$\mathbf{X}_{m,n}$ with respect to tag $t$. $M_n$ is the
total the number of images in the $n$th restaurant. Thus, following the
discussion above, we can use the $k$-largest elements of $\mathbf{G}_n^t$ as the
feature vector of the $n$th restaurant on tag $t$. We denote
such feature vector as $\mathbf{F}_n^t$. Then, for each tag $t$, we can train a
binary classifier $C^t$ using any linear discriminative model, such as SVM.
The procedure of the algorithm is given in Algorithm \ref{alg: restaurant profiling}.
Here, $L$ is the total number of the tags and $N$ is the total number of
restaurants in the training set.

\begin{algorithm}[h]
  \caption{Restaurant profiling algorithm}\label{alg: restaurant profiling}
  \begin{algorithmic}[1]
    \Procedure{Restaurant Profiling}{$k$}
    \For{$n \in \{1, 2, \dots, N\}$}
      \For{$m \in \{1, 2, \dots, M_n\}$}
        \State $\mathbf{S}_{m,n} \gets$ compute the score vector of image $\mathbf{X}_{m,n}$
      \EndFor
    \EndFor
    \For{$t \in \{1, 2, \dots, L\}$}
    \For{$n \in \{1, 2, \dots, N\}$}
      \State $\mathbf{G}_n^t \gets [s_{1,n}^t, s_{2,n}^t, \dots, s_{M_n,n}^t]^T$
      \State $\mathbf{F}_n^t \gets$ $k$-largest elements of $\mathbf{G}_n^t$
    \EndFor
    \State $\mathbf{F}^t \gets \{\mathbf{F}_1^t, \mathbf{F}_2^t, \dots, \mathbf{F}_N^t\}$
    \State $C^t \gets $ train a binary classifier using $\mathbf{F}^t$
    \EndFor
    \State $\mathbf{C} \gets \{C^1, C^2, \dots, C^L\}$
    \State \textbf{return} $\mathbf{C}$
    \EndProcedure
  \end{algorithmic}
\end{algorithm}

\section{Experimental Results} \label{sec: experimental results}
In our experiments, we choose the restaurants that contains both user contributed
photos and user labeled tags for training and testing. Among the $34,787$
restaurants we tracked, $4,573$ satisfy this requirement. We use 5-fold cross-validation
for training and testing. In the training phase, $113,550$ photos are used to
train the multi-label CNN models. We use the BVLC Caffe deep learning framework \cite{jia2014caffe} to
perform all the CNN training and testing. For hyper parameters, we follow
the standard practice on ImageNet challenges\cite{ILSVRC15}, i.e., batch
size $=256$, learning rate $=0.001$, momentum $=0.9$ and weight decay $=5e-06$.
The network is fine tuned from the Alexnet \cite{krizhevsky2012imagenet}, which
was the early winning entry of the ImageNet challenges and considered adequate for this study (although other network structures may provide marginal improvements). The optimization converges after $5$ epochs.
GPUs are used to accelerate our experiments.

\begin{figure*}[h!]
\begin{subfigure}{\textwidth}
  \centering
  \includegraphics[scale=0.45]{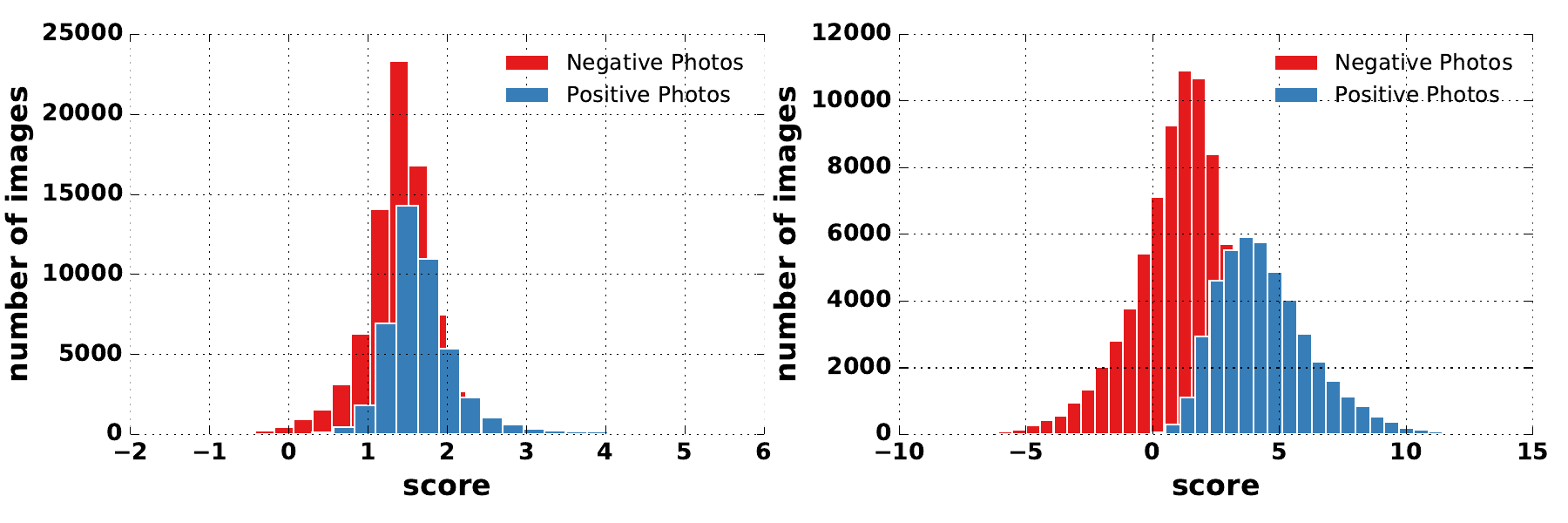}
  \caption{Bar scene}
\end{subfigure}
\begin{subfigure}{\textwidth}
  \centering
  \includegraphics[scale=0.45]{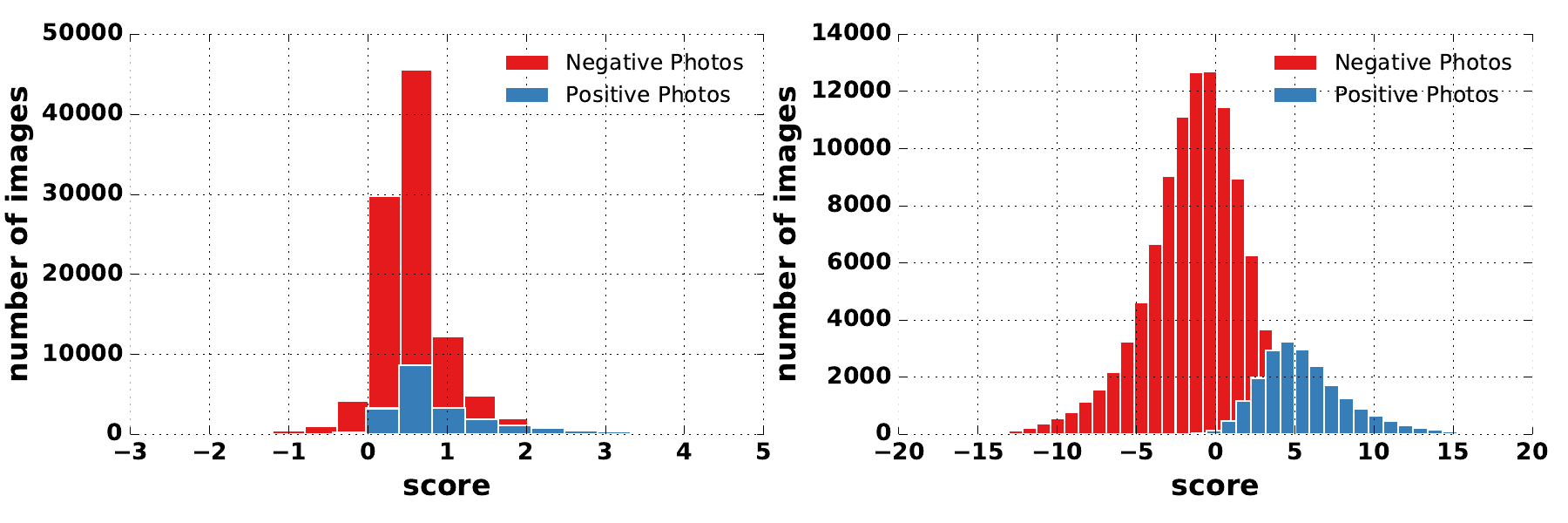}
  \caption{Local cuisine}
\end{subfigure}
\caption{Distributions of image scores on the ``bar scene'' and ``local cuisine''
tags. First column: image scores computed from the first round multi-label CNN.
Second column: image scores computed from the second round multi-label CNN. Negative
photos are those photos without the ``bar scene''/``local cuisine'' tag. Positive
photos are those photos tagged with ``bar scene''/``local cuisine'' tags.}
\label{fig: hits misses distribution}
\end{figure*}

\subsection{Performance of Multi-label CNN} \label{sec: perf multi-label cnn}
For an input image, the multi-label CNN model will output a vector of scores
where each element as the score of the corresponding tag. Ideally, for an
image that is strongly related to the style tag $t$, our multi-label CNN
should output a relatively high score on $t$. On the other hand, if an image
is irrelevant to the tag $t$, then its corresponding score from CNN should
be relatively low. Figure \ref{fig: top scores} shows some top scored images
of different restaurant styles. Not surprisingly, photos with a higher score on
the ``family with children'' style mostly contain
fast foods or snacks, which are very popular among children. Those photos that
receive the top scores on the ``romantic'' style usually contain delicate dishes or
desserts, which are favored by lovers. Finally, for the ``bar scene'' style, top scored images
are often dark with neon lights and contain people drinking wine or beer. Top
scored photos in other rows also show strong relations to their corresponding
tags. It is also very interesting to observe that the top scored images of ``romantic'',
``business meetings'', and ``special occasions'' styles are very similar. It is
reasonable as a restaurant, which is good for ``special occasions'', is usually also
a good place for ``romantic'' scenes or ``business meetings''.

There are also some interesting mistakes. The last image of the second row just contains
some fruits. However, our multi-label CNN gives it a relative high score on the 
``romantic'' tag. The reason for this situation is that in most of the cases, a
``romantic'' photo contains some delicate food that placed in a big white plate.
It looks like the multi-label CNN has learned that in general
a photo with ``romantic'' tag should have a white background. Hence, the fruit
image gets a high score on the ``romantic'' tag. Another mistake can be found in
the last two images of the ``bar scene'' row. Apparently, these two images
only contain the logos of the restaurants and are not related to the ``bar scene''.
The reason for this mistake is that these two restaurants are ``bar scene''
restaurants and  lots of photos contain the logos of these two restaurants
in our training set. Therefore, the logos are learned by the multi-label CNN to the extent that 
images containing such logos will get a high score on the ``bar scene'' tag.

To further evaluate the performance of the multi-label CNN, we use the
two conditional probability equations discussed in Section \ref{sec: pseudo tagging}.
The basic idea of using these two equations is if we see an image with a very
high score on a certain tag $t$, then it is very possible that the image's
corresponding restaurant has a style $t$. Similarly, if an image get a very low
score on tag $t$, then it is unlikely that the corresponding restaurant has a
style $t$. For Equation (\ref{equ: posterior1}), we compute $P(A_t | x > s)$
by counting the proportion of the images whose restaurant has a style $t$
among all the images with a score greater than $s$. For different values of $s$,
we calculate the conditional probability $P(A_t | x > s)$ and the final
probability distribution is given in Figure \ref{fig: posterior}. We can see
from the left figure that when the score $x$ on a tag $t$ is high, the probability
$P(A_t | x > s)$ is also very high. This observation is consistent with our assumption.
A similar analysis can also be applied to Equation (\ref{equ: posterior2}). Therefore,
we can conclude that the trained multi-label CNN performs well as expected on our dataset.

\begin{table*}[h!]
  \centering
    \begin{tabular}{l|c|c|c|c|c|c}
      \hline
      \multicolumn{1}{c}{\multirow{2}{*}{\textbf{Styles}}} & \multicolumn{3}{|c|}{\textbf{Baseline}} &  \multicolumn{3}{c}{\textbf{Proposed Method}}      \\ \cline{2-7}
                               & \textbf{Rec(\%)} & \textbf{Prec(\%)} & \textbf{F-1 score(\%)} &  \textbf{Rec(\%)} & \textbf{Prec(\%)} & \textbf{F-1 score(\%)} \\
      \hline
      Bar scene                &  \textbf{100.00} &           36.69   &              53.68 &             80.39 &    \textbf{48.81} &     \textbf{60.74} \\
      Business meeting         &            37.21 & \textbf{84.21}    &              51.61 &    \textbf{76.74} &             52.38 &     \textbf{62.26} \\
      Dining on a budget       &            11.76 & \textbf{100.00}   &              21.05 &    \textbf{94.12} &             35.56 &     \textbf{51.61} \\
      Families with children   &  \textbf{100.00} &           67.15   &              80.35 &             77.17 &    \textbf{86.59} &     \textbf{81.61} \\
      Large groups             &   \textbf{71.67} &           39.45   &              50.89 &             58.33 &    \textbf{50.72} &     \textbf{54.26} \\
      Local cuisine            &             3.85 &           33.33   &               6.90 &    \textbf{69.23} &    \textbf{48.65} &     \textbf{57.14} \\
      Romantic                 &            17.24 & \textbf{100.00}   &              29.41 &    \textbf{89.66} &             35.14 &     \textbf{50.49} \\
      Scenic view              &            14.29 & \textbf{100.00}   &              25.00 &    \textbf{71.43} &             17.86 &     \textbf{28.57} \\
      Special occasions        &            37.04 &  \textbf{80.00}   &              50.63 &    \textbf{90.74} &             68.06 &     \textbf{77.78} \\
      \hline
      \textbf{Average}         &            43.67 &    \textbf{71.20} &              41.06 &    \textbf{78.65} &             49.31 &      \textbf{58.27}\\
      \hline
    \end{tabular}
  \caption{Performance of the baseline method and the proposed method on different style tags. Here, we choose $k'=3$ and
  the minimum number of photos is $50$. Note that the F-1 scores have improved across all the styles due to the proposed method. }
  \label{tab: performance}
\end{table*}

\subsection{Performance of Pseudo Tagging}
As we have discussed in Section \ref{sec: perf multi-label cnn}, the photo scores
output by the first round multi-label CNN are good indicators of the style
tags. As shown in Figure \ref{fig: posterior}, the first round multi-label
CNN performs very well when the score is very high or very low. However, when
the score of an image falls in the middle, it is  difficult to tell whether it
belongs to the style tag or not.

Figure \ref{fig: hits misses distribution} shows the score distributions before
and after using the pseudo tagging algorithm. Here, the first row shows the
image score distributions of the ``bar scene'' tag. The second row shows the
image score distributions of the ``local cuisine'' tag. The figures in the first 
column show the distributions of image scores computed from the 
first round multi-label CNN. For the ``bar scene'' tag, most of
the images have a score between $1$ and $2$ while the positive and
negative images are mixed together. It means for images with scores between $1$
and $2$, the first round multi-label CNN model cannot distinguish their
styles. This is also the case for the ``local cuisine'' tag.
After applying the pseudo tagging algorithm to our training set, we
train the multi-label CNN for the second round and the distributions of the computed
image scores are shown in the second column of Figure \ref{fig: hits misses distribution}.
We can find that in this case negative and positive images do not mix as 
much as before, which means the second round multi-label CNN has a better performance in
distinguishing the styles of images. Therefore, the binary classifier training
in the restaurant profiling algorithm can benefit significantly from the better
separated score distributions. We can also see that our pseudo tagging algorithm
helps in separating the score distribution of the ``local cuisine'' tag. That is 
why there is a much improved performance gain for the ``local cuisine'' tag in Table
\ref{tab: performance}.

\subsection{Overall Performance} \label{sec: overall perf}
To show the performance gain from the pseudo tagging and the restaurant profiling
algorithm, we also establish a baseline for comparison. For the baseline method, we only
train the multi-label CNN once and assign top-$k'$ scored tags to photos. If more
than a half of the photos are assigned a particular tag, the restaurant is determined to be labeled as such.

In our experiments, we test the performance of the baseline method and our
proposed method on a variety of choices of $k'$ and $k$. We notice that
the more photos a restaurant have, the more information about the restaurant's styles
we can extract. Therefore, we also set different minimum numbers of photos to restaurants
in the test set. The performances of the two methods with the best parameter
settings are given in Table \ref{tab: performance}. Following the evaluation
methods in \cite{BMVC.25.59,Zhou20122291}, we use recall, precision, and F-1 measure 
as the metrics to evaluate the performances of the two methods on different style tags.
We can see from Table \ref{tab: performance}
that, in general, the proposed method performs much better than the baseline method.
It means that the pseudo tagging and the restaurant profiling algorithms play
a key role in improving the performance of the proposed method. An interesting
finding here is that the ``scenic view'' style has a very low precision. A closer
look at the dataset finds that there are a number of the restaurants contain
outdoor images. But those restaurants cannot be considered as typical ``scenic view''
style restaurant and are not labeled with the ``scenic view'' tag.

\begin{figure}
  \centering
  \includegraphics[scale=0.28]{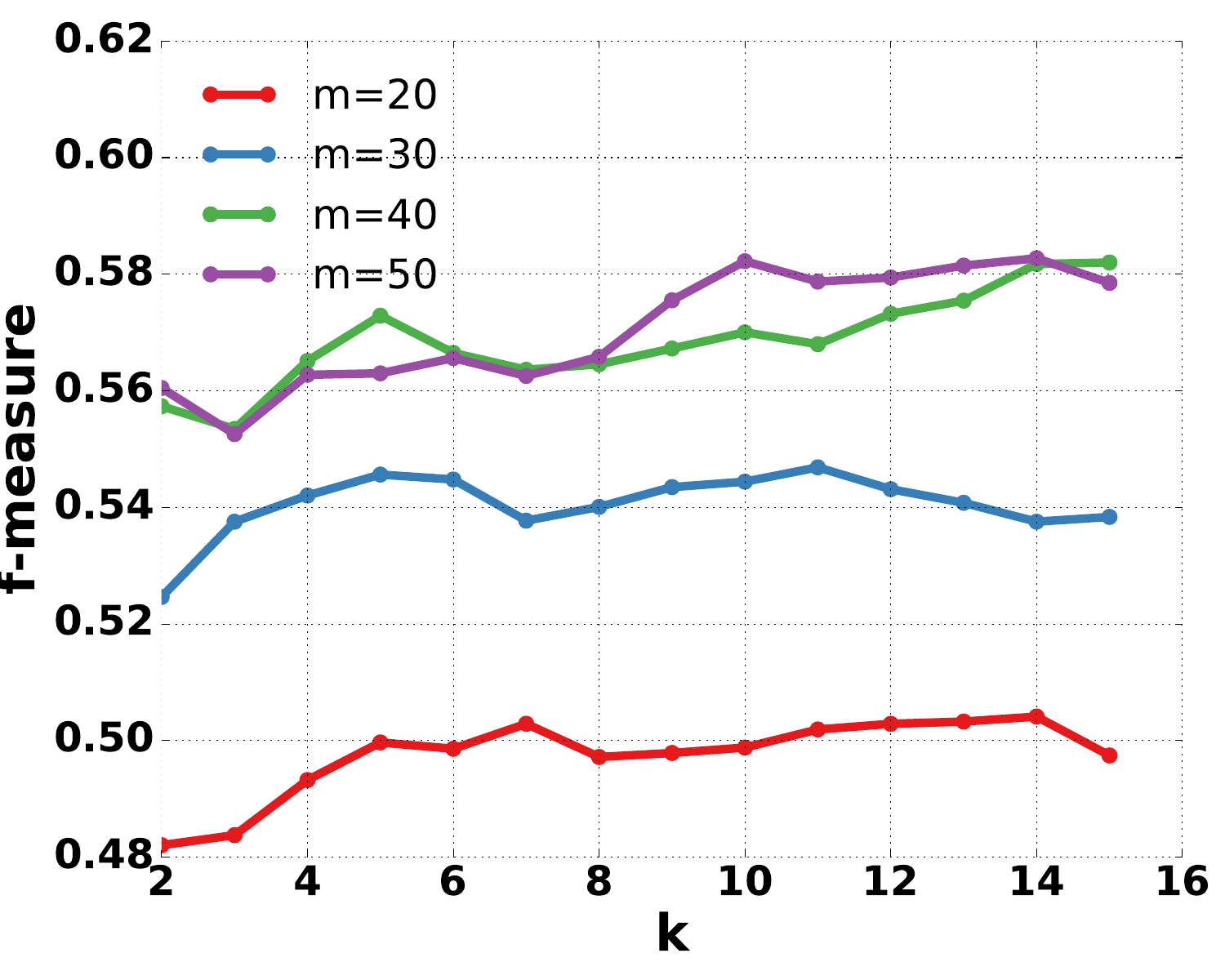}
  \caption{The average f-measure scores of our method when choosing different
  $k$ and minimum numbers of photos.}
  \label{fig: k vs minimum photos}
\end{figure}

We also investigate the performance differences of our proposed method when
choosing different parameter settings. From Figure \ref{fig: k vs minimum photos},
we can find that the choice of $k$ does not affect the performance  much.
However, when we choose the restaurants with a higher minimum number of photos
the performance gets better. It means when the number of photos of a restaurant
is high, our proposed method can achive more accurate estimates of the
restaurant styles.

\section{Conclusions} \label{sec: conclusions}
We have presented a novel approach to profiling restaurant styles directly from user
uploaded photos on user-review sites. We propose to build a deep MIML framework
to deal with the special problem setting of restaurant style classification.
Due to the absence of individual photo tags, we initially
train the multi-label CNN using photos labeled with all the restaurant tags supplied by users. We then
refine the photo tags using our pseudo tagging algorithm and train the multi-label
CNN for a second round. Experiments show that the multi-label CNN performs very
well in inferring the restaurant styles, and the pseudo tagging algorithm plays a key role in
helping the multi-label CNN to distinguish different restaurant styles. Finally, the photo-level style
estimates are used by the restaurant profiling algorithm to train a binary classifier
for each style tags using SVM. Our experimental results show that our approach has
achieved a significant performance gain due to the pseudo tagging and restaurant profiling
algorithms. We also show that when the number of photos of a restaurant increases,
the performance of our approach increases as well.

\section*{Acknowledgment}
We gratefully acknowledge the support from the University, New York
State through the Goergen Institute for Data Science, and our
corporate sponsors Xerox and Yahoo.



\bibliographystyle{IEEEtran}
\bibliography{reference}
%

\end{document}